\documentclass{article}
\usepackage{spconf,amsmath,graphicx}
\usepackage{xcolor}
\usepackage{enumitem}
\usepackage{cancel}
\usepackage{makecell}
\usepackage{multirow, multicol}
\usepackage{threeparttable}
\usepackage{balance}
\usepackage{bbding}
\usepackage{caption}
\usepackage{threeparttable}
\usepackage{bm}
\usepackage{booktabs}

\title{LEARN FROM ZOOM: DECOUPLED SUPERVISED CONTRASTIVE LEARNING FOR WCE IMAGE CLASSIFICATION}
\name{Kunpeng Qiu$^{1,2}$, Zhiying Zhou$^{1,2}$,Yongxin Guo$^{1,2}$}
\address{$^1$Department of Electrical and Computer Engineering, National University of Singapore, Singapore \\
$^2$National University of Singapore Suzhou Research Institute, China}
\begin{document}
%
\maketitle
\begin{abstract}

Accurate lesion classification in Wireless Capsule Endoscopy (WCE) images is vital for early diagnosis and treatment of gastrointestinal (GI) cancers. However, this task is confronted with challenges like tiny lesions and background interference. Additionally, WCE images exhibit higher intra-class variance and inter-class similarities, adding complexity. To tackle these challenges, we propose \textit{Decoupled Supervised Contrastive Learning} for WCE image classification, learning robust representations from zoomed-in WCE images generated by \textit{Saliency Augmentor}. Specifically, We use uniformly down-sampled WCE images as anchors and WCE images from the same class, especially their zoomed-in images, as positives. This approach empowers the \textit{Feature Extractor} to capture rich representations from various views of the same image, facilitated by \textit{Decoupled Supervised Contrastive Learning}. Training a linear \textit{Classifier} on these representations within 10 epochs yields an impressive \textbf{92.01\%} overall accuracy, surpassing the prior state-of-the-art (SOTA) by \textbf{0.72\%} on a blend of two publicly accessible WCE datasets. Code is available at:  \texttt{https://github.com/Qiukunpeng/DSCL}.


\end{abstract}
\begin{keywords}
Wireless Capsule Endoscopy, Lesion classification, Saliency Augmentor, Contrastive Learning
\end{keywords}
\section{Introduction}
\label{sec:intro}

Accurate identification and classification of vascular lesions and inflammation in WCE images are crucial for early diagnoses of GI abnormalities such as bleeding, ulcers, and Crohn's disease \cite{georgakopoulos2016weakly}. Despite significant advances in deep learning \cite{litjens2017survey}, automatically identifying these conditions in WCE images remains challenging. The curse of dimensionality \cite{berisha2021digital} caused by limited WCE annotation samples and tiny lesion areas leads to overfitting problem \cite{guo2020semi}.



To tackle this challenge, numerous approaches have been proposed, including transfer learning \cite{shang2019leveraging}, dropout \cite{srivastava2014dropout}, mixup \cite{zhang2017mixup}, and label-smoothing regularization \cite{muller2019does}. Another effective solution is saliency-based attention, where a CNN naturally identifies task-salient regions \cite{zhou2016learning}, encouraging the network to focus more on these areas. Among saliency-based attention methods, Recasens et al. \cite{recasens2018learning} used saliency maps to zoom in task-salient regions to help the classification network capture discriminative features. Xing et al. \cite{8759401} created saliency-aware inputs to highlight lesion regions. They later proposed a dual attention model \cite{9143178} to enhance lesion recognition by combining zoomed-in lesion features with original ones. Guo and Yuan \cite{guo2019triple} incorporated a trainable abnormal-aware attention module to improve abnormality detection. Additionally, George et al. \cite{10095510} suggested aggregating saliency maps with RGB images to enhance WCE image classification. 

While saliency-based attention mechanisms can alleviate overfitting due to curse of dimensionality by enhancing lesion features and suppressing irrelevant background features, they struggle to effectively address intra-class variance and inter-class similarities using cross-entropy loss. The cross-entropy loss is typically computed for an individual sample, which doesn't inherently capture the relationship between the samples in a batch \cite{graf2021dissecting,khosla2020supervised, zou-etal-2023-unis}. Furthermore, these methods depend on saliency images for resampling the original images, which is a process known to be highly time-consuming, making such networks unsuitable of practical deployment.

\begin{figure*}[ht]
\centering
\includegraphics[width=0.9\textwidth]{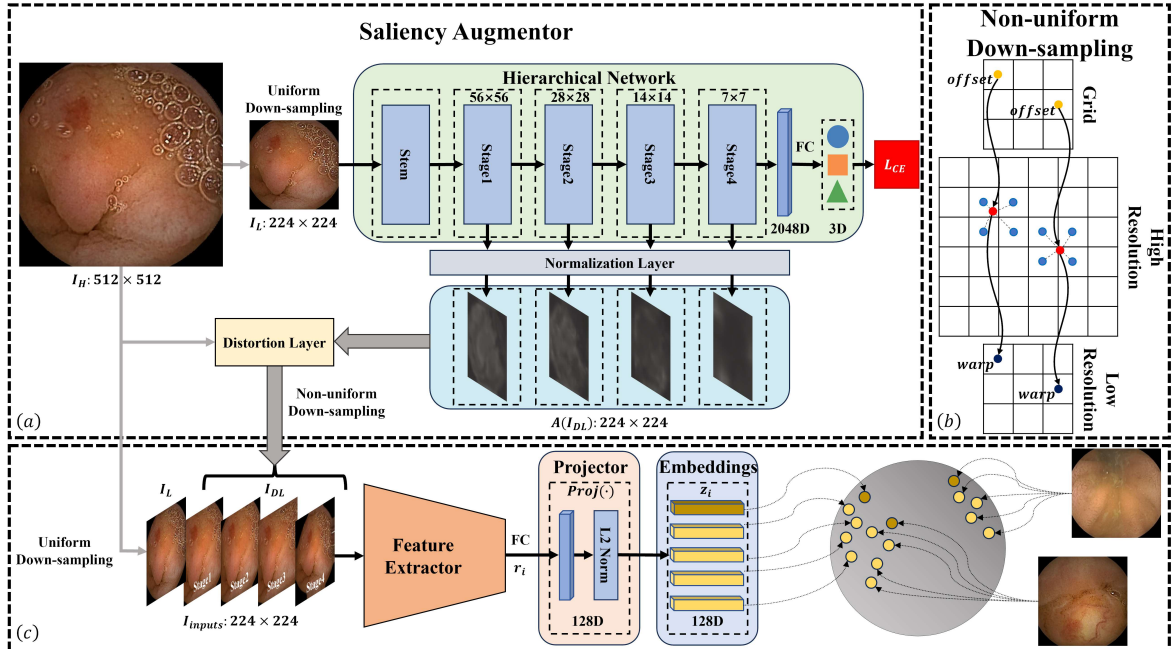}
\caption{\small The overall architecture of our proposed method. (a) Framework of Saliency Augmentor; (b) Principle diagram of non-uniform down-sampling, and (c) Framework of Decoupled Supervised Contrastive Learning.}
\vspace{-0.5cm}
\label{fig: fig02}
\end{figure*}

In this paper, we propose a novel contrastive learning approach for WCE image classification based on a saliency-driven attention mechanism to overcome the aforementioned challenges. Contrastive learning in computer vision heavily relies on data augmentation \cite{chen2020simple}, which is a technique widely explored and applied with the ImageNet dataset \cite{russakovsky2015imagenet} in SimCLR \cite{chen2020simple}. However, these strategies are not task-agnostic, especially for WCE image classification. Inspired by previous methods \cite{recasens2018learning,8759401,9143178,guo2019triple,10095510}, we abstain from utilizing zoomed-in WCE images as the primary training data for the task network. Instead, we use uniformly down-sampled WCE images as anchors and WCE images from the same class, especially their zoomed-in images, as positives. We construct diverse contrastive tuples by incorporating multiple views of the same image at different stages, thereby diversifying the input combinations \cite{tian2020contrastive}. This process enhances embeddings for both intra-class compactness and inter-class separability \cite{graf2021dissecting,khosla2020supervised,10096794}. Additionally, we propose a novel \textit{Decoupled Supervised Contrastive Learning} loss to facilitate convergence.

The main contributions are summarized as follows: 
(1) We propose the \textit{Decoupled Supervised Contrastive Learning}, effectively enhancing intra-class similarity and inter-class variance in the feature distribution of the task model. Due to the decoupling, the task network demonstrates more stable and rapid convergence. 
(2) To extract more robust and fine-grained WCE lesion features, we propose the \textit{Saliency Augmentor}. Unlike direct training on the zoomed-in images, our method employs uniformly down-sampled WCE images as anchors and images from the same class, especially their zoomed-in images, as positives, ensuring greater stability and avoiding the time-consuming resampling process during deployment. 
(3) Our experimental results, conducted on a blend of two publicly available WCE datasets, demonstrated the effectiveness and superiority of our proposed method.

\section{PROPOSED METHOD}
\label{sec:format}

Following the common contrastive learning training paradigm, our approach consists of two stages. In the first stage, the uniformly down-sampled image $I_L$ and the non-uniformly down-sampled images $I_{DL}$ generated by the \textit{Saliency Augmentor} from the same WCE image are combined as inputs $I_{inputs}$ for the \textit{Feature Extractor}. A linear \textit{Projector} is employed to map the 2048-dimensional output from the global average pooling layer into a reduced 128-dimensional space. This \textit{Feature Extractor} is trained using \textit{Decoupled Supervised Contrastive Learning} loss to develop distinctive features. In the second stage, the \textit{Saliency Augmentor} and \textit{Projector} are discarded while keeping the parameters of the \textit{Feature Extractor} frozen. A linear \textit{Classifier} is trained using the cross-entropy loss. An overview of our approach is visually depicted in Fig. 1.


\setlength{\tabcolsep}{2.3mm}{
\begin{table*}[t]
    \centering
    \begin{threeparttable}
    \caption{Comparison with SOTA methods for classification of WCE images.}\label{tab1}
    \setlength{\tabcolsep}{3mm}
    \begin{tabular}{ c| c c c c c| c}
        \toprule[2pt]
        Methods & N-Rec (\%) & V-Rec (\%) &  I-Rec (\%) &  OA (\%) & CK (\%) & IT (ms/image)\\ 
        \midrule[0.75pt]
        He et al. \cite{he2016deep} & 93.98$\pm$0.58 & 78.90$\pm$1.65 & 81.78$\pm$0.89 & 86.20$\pm$0.36 & 78.74$\pm$0.53 & \textbf{0.39} \\
        Recasens et al. \cite{recasens2018learning} & 96.09$\pm$0.97 & 81.32$\pm$1.20 & 86.78$\pm$0.70 & 89.19$\pm$0.30 & 83.35$\pm$0.47 & 0.73 \\
        Guo et al. \cite{guo2019triple} & 95.79$\pm$0.60 & 89.50$\pm$0.53 & 84.41$\pm$1.35 & 89.90$\pm$0.31 & 84.85$\pm$0.46 & 5.71\tnote{$^{*}$} \\		
        Xing et al. \cite{9143178} & 95.72$\pm$0.65 & \textbf{90.72$\pm$0.70} & 87.44$\pm$1.70 & 91.29$\pm$0.35 & 86.97$\pm$0.52 & 4.22\tnote{$^{*}$}  \\
        \midrule[0.75pt]
        Our method & \textbf{96.46$\pm$0.51} & 88.90$\pm$1.53 & \textbf{88.33$\pm$0.29} & \textbf{92.01$\pm$0.45} & \textbf{87.73$\pm$0.70} & \textbf{0.39} \\
        \bottomrule[2pt]
    \end{tabular}
    \begin{tablenotes}
      \item[$^{*}$ is implemented by us.]
    \end{tablenotes}
    \end{threeparttable}
    \vspace{-0.5cm}
\end{table*}}

\subsection{Saliency Augmentor (SA)}
\label{ssec:subhead}
As shown in Fig. 1(a), a high-resolution WCE image $I_H$ is initially uniformly down-sampled to $224 \times 224$ resolution to create $I_L$. It is then processed by a hierarchical network, yielding feature maps at different stages. These feature maps are condensed into a single-layer feature map $A(I_{DL})$ using $1 \times 1$ convolution, followed by softmax normalization. Similar to \cite{recasens2018learning}, a distance kernel $k((x, y), (x^{\prime}, y^{\prime}))$ is employed to generate a saliency map. This map guides the non-uniform down-sampling of $I_H$ into $I_{DL}$, emphasizing the lesion area while compressing background noise. And the non-uniform down-sampling procedure is represented as:
\begin{equation}
    \setlength{\abovedisplayskip}{2pt}
    \setlength{\belowdisplayskip}{2pt}
    \mathcal{T} : (x, y) \rightarrow (x', y')
\end{equation}
\begin{equation}
    \setlength{\abovedisplayskip}{2pt}
    \setlength{\belowdisplayskip}{2pt}
    I'(x, y) = I(\mathcal{T}^{-1}(x, y))
\end{equation}

\noindent
where $(x, y)$ and $(x^{\prime}, y^{\prime})$ denote coordinates in $I_H$ and $I_{DL}$. As illustrated in Fig. 1(b), each grid position within the $I_{DL}$ undergoes a backward mapping operation, calculating its inverse mapping $\mathcal{T}^{-1}$ to establish corresponding coordinates in $I_H$. Essentially, the value of $I_{DL}$ is determined through bilinear interpolation from neighboring pixels in $I_H$, with neighborhoods defined by the $offset$ in a learned grid field.

Given that non-uniform down-sampling is a discrete process, and different stages of the hierarchical network emphasize distinct lesion features, such as edges, colors, and various lesion regions, we leverage the four feature maps generated by the network to perform individual non-uniform down-sampling on $I_H$. This procedure enables the creation of multiple views of the same image. Considering the diminutive size of lesions in WCE images, we introduce an offset temperature hyperparameter, denoted as $\tau_o$ (where $\tau_o$ is less than 1), into the softmax normalization process. This inclusion enhances grid offset, effectively reducing background noise.

\subsection{Decoupled Supervised Contrastive Learning (DSCL)}
\label{ssec:subhead}

Supervised Contrastive Learning (SCL) \cite{khosla2020supervised} has demonstrated remarkable performance by incorporating label information. In Fig. 1(c), the inputs include both the uniformly down-sampled anchor $I_L$ and the non-uniformly down-sampled positives $I_{DL}$, generated by the \textit{SA} from the same WCE image. These inputs $I_{inputs}$ are processed by the \textit{Feature Extractor}, producing feature embeddings $r_i \in R^{D}$. Subsequently, these embeddings are projected to $z_i \in R^{d}$ ($d < D$) through $Proj(\cdot)$. The embeddings $z_i$ are then L2 normalized to lie on the unit hypersphere, enabling similarity measurement via inner product. In the unit hypersphere, SCL treats samples of the same class as positive samples, not just data augmentation of anchor, encouraging their representations to get closer, while treating images from different classes as negatives, pushing their representations apart. 

However, similar to self-supervised contrastive learning \cite{he2020momentum}, SCL exhibits a negative-positive coupling (NPC) effect, which necessitates substantial computational resources to ensure efficient learning. Motivated by \cite{yeh2022decoupled}, we address the NPC effect in SCL by eliminating the positive term from the loss denominator, resulting in the \textit{DSCL} loss. Finally, for each model sample $z_i$, we define the \textit{DSCL} loss as follows:
\begin{equation}
    \setlength{\abovedisplayskip}{2pt}
    \setlength{\belowdisplayskip}{2pt}
    \mathcal{L}_{DSCL} = - \frac{1}{P} \sum_{p=1}^{P} \log \frac{e^{(z_i \cdot z_p / \tau)}}{\bcancel{e^{(z_i \cdot z_p / \tau)}} + \sum_{a \in A(i)} e^{(z_i \cdot z_a / \tau)}}
\end{equation}
\noindent
where $\tau$ controls the concentration level, $i$ represents the anchor index, $p$ is the positive sample index (distinct from $i$), $P$ is the total number of positive samples, and $A(i)$ is the set containing all samples except the anchor. The positive term is removed from the loss denominator. As suggested in \cite{khosla2020supervised}, the summation over positives is placed outside the $\log$.

\subsection{Training and Testing}
\label{ssec:subhead}

In the proposed framework, during the first stage, the \textit{SA} is optimized using the cross-entropy loss function, while the \textit{Feature Extractor} is optimized using our proposed \textit{DSCL} loss. The final optimization objective is as follows:
\begin{equation}
    \setlength{\abovedisplayskip}{2pt}
    \setlength{\belowdisplayskip}{2pt}
    \mathcal{L}_{S1} = \mathcal{L}_{CE} + \mathcal{L}_{DSCL}
\end{equation}
During the second stage, we discard the \textit{SA} and \textit{Projector}, while keeping the parameters of the \textit{Feature Extractor} frozen. A linear \textit{Classifier} is trained using the cross-entropy loss function. The final optimization objective is as follows:
\begin{equation}
    \setlength{\abovedisplayskip}{2pt}
    \setlength{\belowdisplayskip}{2pt}
    \mathcal{L}_{S2} = \mathcal{L}_{CE}
\end{equation}
At the testing stage, similar to the second stage, we omit the \textit{SA} and \textit{Projector}, resulting in an inference time almost equivalent to the vanilla \textit{Feature Extractor}.

\section{EXPERIMENTS}
\label{sec:pagestyle}

\subsection{Dataset}
\label{ssec:subhead}

We evaluated our method on a combined dataset of 3022 images, merging CAD-CAP \cite{leenhardtCADCAP25000image2020} (1812 images) and KID \cite{koulaouzidisKIDProjectInternetbased2017} (1210 images) datasets. The dataset includes three classes: normal images (1300 images), vascular lesions (888 images), and inflammatory lesions (834 images). Images were standardized to 512×512 resolution, borders were removed, and data augmentation (flipping and croping) ensured robustness. We use the 5-fold cross-validation strategy to validate the effectiveness and robustness of the proposed method.

\subsection{Implementation Details}
\label{ssec:subhead}

\textbf{Backbone Architecture:} We leverage the ResNet50 architecture \cite{he2016deep} for both \textit{SA} and the \textit{Feature Extractor}.

\noindent
\textbf{Network Training:} Our approach consists of two training stages. In the first stage, we trained the model for 200 epochs using $\mathcal{L}_{S1}$. We employed the SGD optimizer with Nesterov momentum and a batch size of 32. The initial learning rate for \textit{SA} was set to 1e-1, and for the \textit{Feature Extractor}, it was 1e-2, following a cosine annealing strategy, both with a weight decay of 5e-4. We set $\tau_o$ to 0.1 and $\tau$ to 0.07. In the subsequent stage, we used $\mathcal{L}_{S2}$ to train the linear \textit{Classifier} for 10 epochs, excluding \textit{SA} and \textit{Projector} from this phase. All other settings remained consistent with the first stage of the \textit{Feature Extractor}.

Our method, implemented in PyTorch, ran on a workstation equipped with an Intel Xeon GOLD 6226R 2.9 GHz processor and an NVIDIA TITAN RTX GPU.
\setlength{\tabcolsep}{2.3mm}{
\begin{table}[t]
    \centering
    \caption{Ablation study on the proposed model.}\label{tab2}
    \setlength{\tabcolsep}{0.75mm}
    \begin{tabular}{ c| c c | c c | c}
        \toprule[2pt]
        \multirow{2}*{Methods} & \multicolumn{2}{c|}{Augmentation} & \multicolumn{2}{c|}{Loss} & \multirow{2}*{OA (\%)} \\
        ~ & SimCLR \cite{chen2020simple} & SA &  $\mathcal{L}_{SCL}$ & $\mathcal{L}_{DSCL}$ & ~ \\
        \midrule[0.75pt]
        Baseline1 & \CheckmarkBold & ~ & \CheckmarkBold & ~ & 65.92$\pm$2.05 \\
        Baseline2 & \CheckmarkBold & ~ & ~ & \CheckmarkBold & 67.17$\pm$0.56 \\
        Baseline3 & ~ & \CheckmarkBold & \CheckmarkBold & ~ & 90.65$\pm$0.35 \\
        \midrule[0.75pt]
        Our method & ~ & \CheckmarkBold & ~ & \CheckmarkBold & \textbf{92.01$\pm$0.45} \\
        \bottomrule[2pt]
    \end{tabular}
    \vspace{-0.5cm}
\end{table}}

\noindent
\textbf{Evaluation Metrics:} We evaluated the performance of all SOTA methods using the following metrics: Recall of Normal Images (N-Rec), Recall of Vascular Lesions (V-Rec), Recall of Inflammatory Images (I-Rec), Overall Accuracy (OA), and Cohen’s Kappa Score (CK). Inference Time (IT) is used to assess the computational efficiency during deployment.

\begin{figure}[htb]
\begin{minipage}[b]{.48\linewidth}
  \centering
  \centerline{\includegraphics[width=4.0cm]{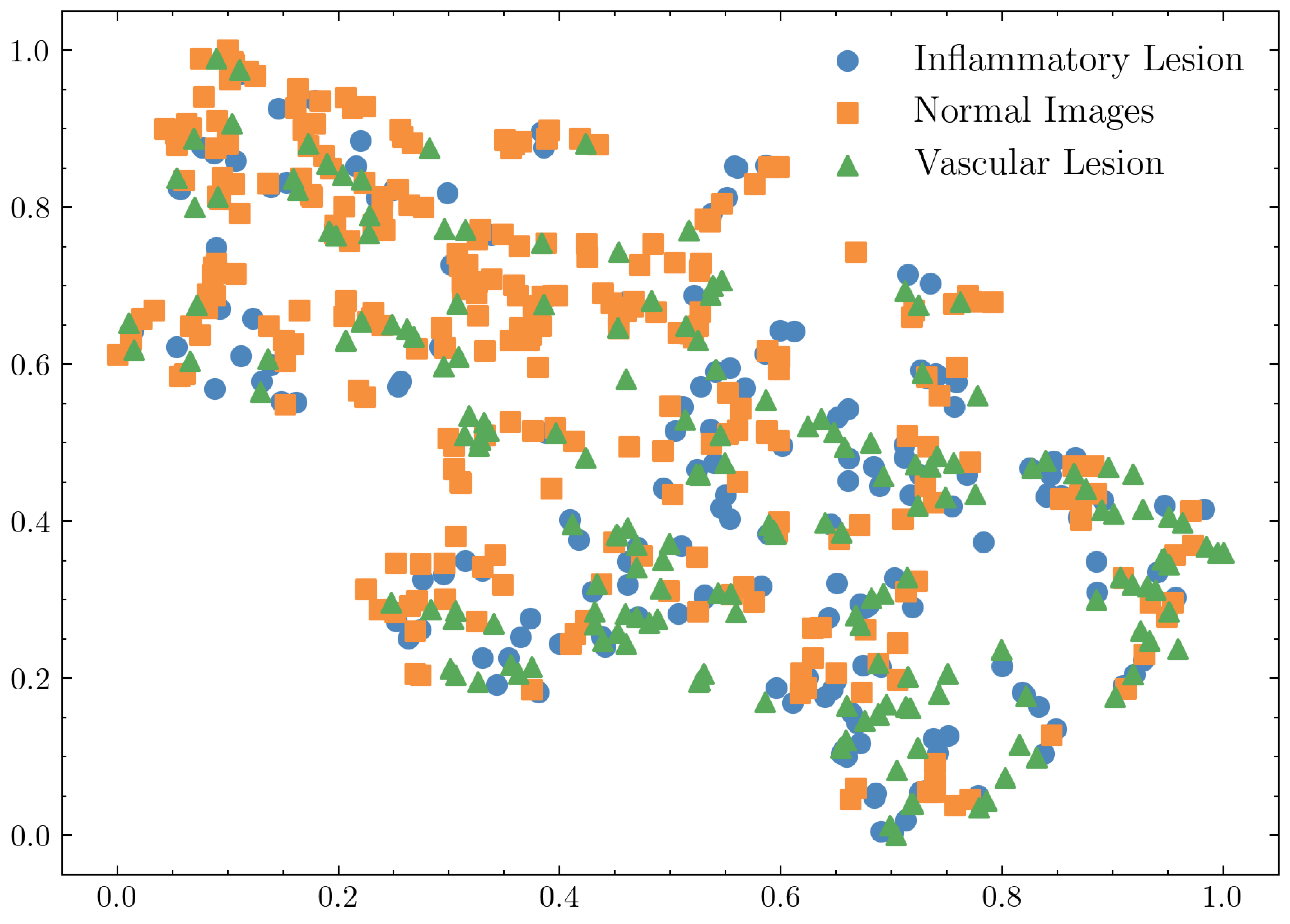}}
  \centerline{(a) Original WCE Images}\medskip
\end{minipage}
\hfill
\begin{minipage}[b]{0.48\linewidth}
  \centering
  \centerline{\includegraphics[width=4.0cm]{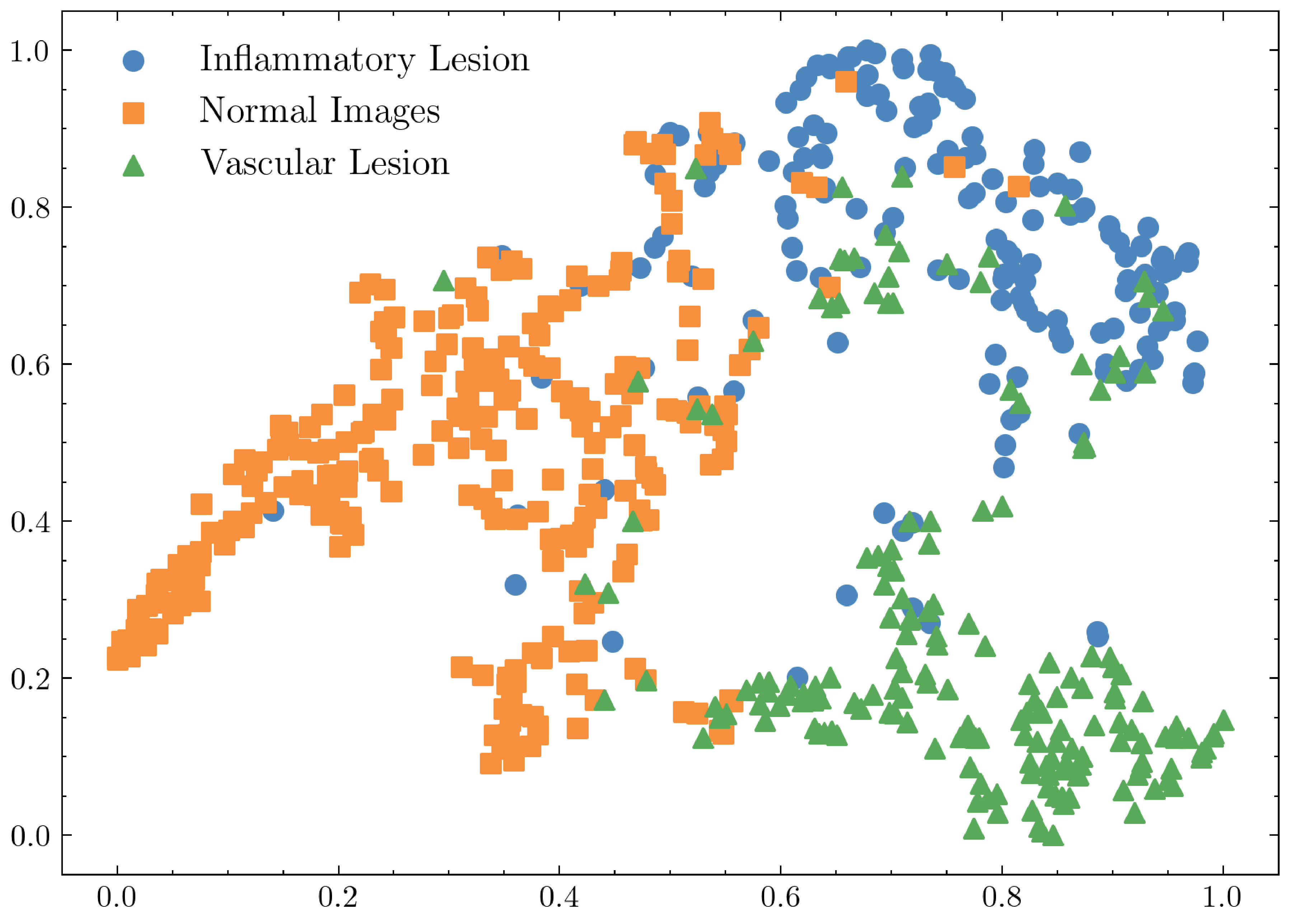}}
  \centerline{(b) ResNet50 w/ $\mathcal{L}_{CE}$}\medskip
\end{minipage}
\hfill
\begin{minipage}[b]{0.48\linewidth}
  \centering
  \centerline{\includegraphics[width=4.0cm]{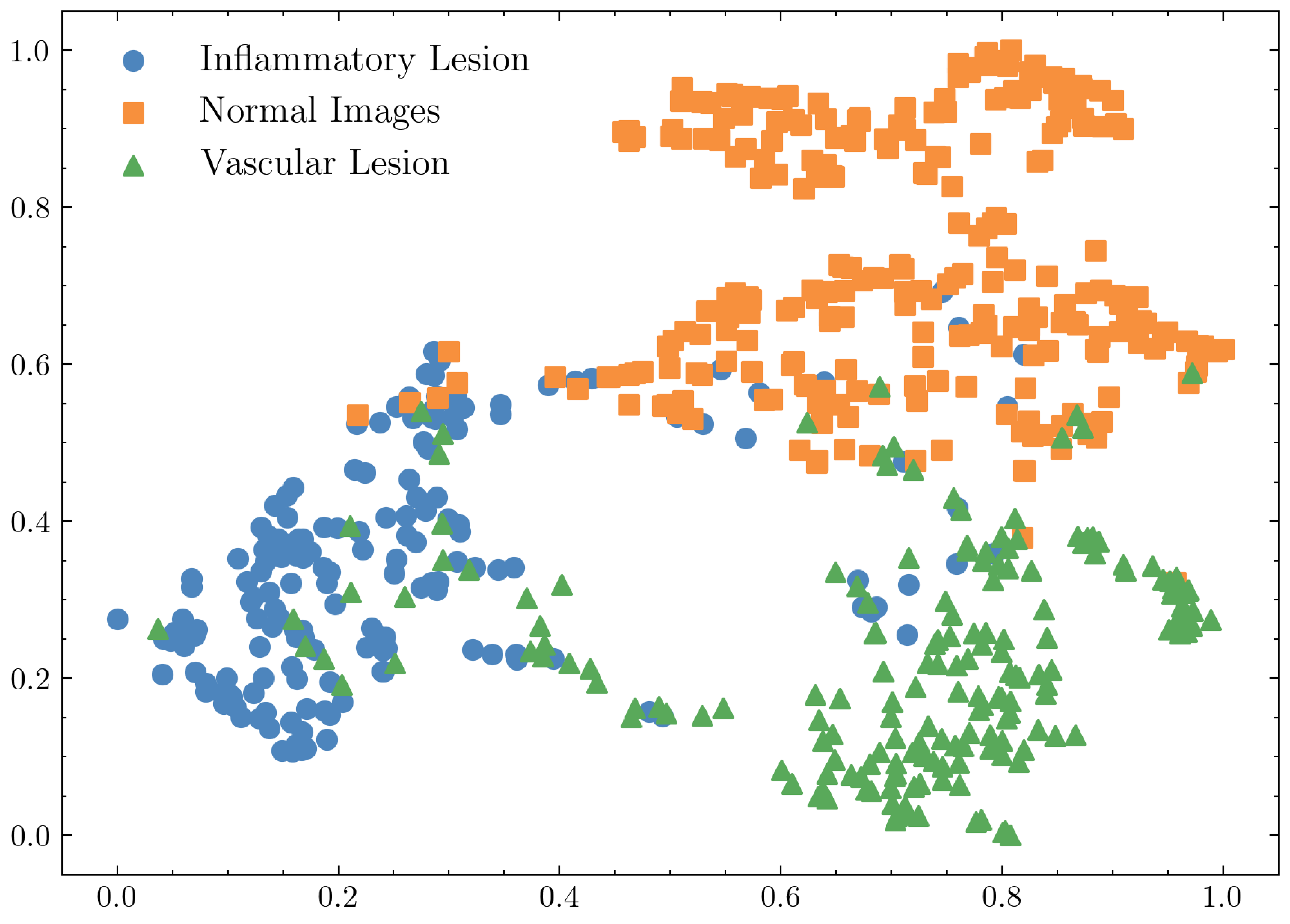}}
  \centerline{(c) SaliencyNet w/ $\mathcal{L}_{CE}$}\medskip
\end{minipage}
\hfill
\begin{minipage}[b]{0.48\linewidth}
  \centering
  \centerline{\includegraphics[width=4.0cm]{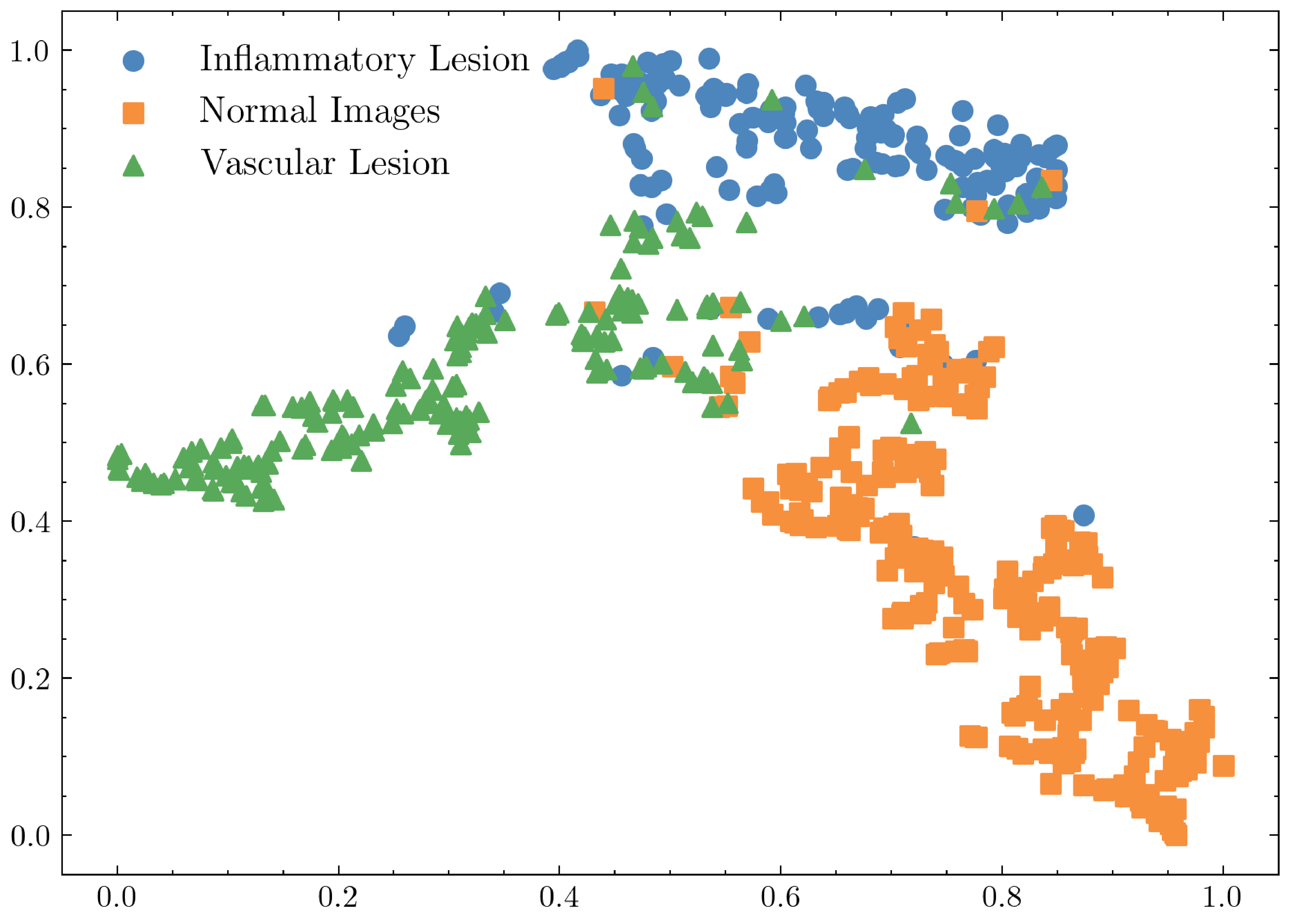}}
  \centerline{(d) ResNet50 w/ $\mathcal{L}_{DSCL}$}\medskip
\end{minipage}

\caption{\small  The t-SNE Visualization of Feature Distribution. (a) Original WCE Images; (b) Output of $\mathcal{L}_{CE}$; (c) Output of $\mathcal{L}_{CE}$ with Zoomed-In; (d) Output of $\mathcal{L}_{DSCL}$ with Zoomed-In.}
\label{fig: fig02}
\vspace{-0.5cm}
\end{figure}

\section{RESULTS AND ANALYSIS}
\label{sec:typestyle}

\subsection{Results and Comparison}
\label{ssec:subhead}

We compared our method with four deep learning-based WCE image classification approaches. The results in Table 1 demonstrate the superiority of our model. In comparison to the SOTA method \cite{9143178}, our approach exhibits significant improvements in N-Rec, I-Rec, OA, and CK, with gains of 0.74\%, 0.89\%, 0.72\%, and 0.76\%, respectively. During the inference stage, our method significantly outperforms existing methods in terms of speed that heavily rely on resampling.

\subsection{Ablation Study}
\label{ssec:subhead}

To analyze the contributions of our proposed method, Table 2 quantitatively presents the performance of Baseline1 and our method with the same \textit{Feature Extractor} ResNet50 \cite{he2016deep}. We conducted additional comparative experiments to further dissect the impact of each component.

\setlength{\tabcolsep}{2.3mm}{
\begin{table}[t]
    \centering
    \begin{threeparttable}
    \caption{Quantitative comparison of $\mathcal{L}_{DSCL}$ and $\mathcal{L}_{CE}$.}\label{tab3}
    \setlength{\tabcolsep}{1.5mm}
    \begin{tabular}{ c| c | c | c}
        \toprule[2pt]
        Methods & Loss & Intra-Class \bm{$\uparrow$} & Inter-Class \bm{$\downarrow$} \\ 
        \midrule[0.75pt]
        ResNet \cite{he2016deep} & $\mathcal{L}_{CE}$ & 0.65 & -0.30 \\
        SaliencyNet\tnote{$^{*}$} \cite{recasens2018learning} & $\mathcal{L}_{CE}$ & 0.70 & -0.32 \\		
        \midrule[0.75pt]
        Our method & $\mathcal{L}_{DSCL}$ & \textbf{0.79} & \textbf{-0.35} \\
        \bottomrule[2pt]
    \end{tabular}
    \begin{tablenotes}
      \item[$^{*}$ is named by us.]
    \end{tablenotes}
    \end{threeparttable}
    \vspace{-0.5cm}
\end{table}}
Consistent with the number of data augmentations in our proposed method, we used five random augmentations following the SimCLR \cite{chen2020simple} data augmentation scheme, resulting in the baseline1 overall accuracy of 65.96\%. Compared to baseline1, introducing \textit{DSCL} improved performance by 1.21\%, demonstrating its effectiveness. Our proposed \textit{SA} significantly enhanced baseline1 to 90.65\%, a substantial 24.69\% increase, highlighting its efficacy for WCE image classification. Combining \textit{DSCL} further improved performance by 1.36\%, leading to our final method.



\subsection{Analysis and visualization}
\label{ssec:subhead}

To assess the effectiveness of $\mathcal{L}_{DSCL}$ in addressing intra-class and inter-class similarity challenges, we conducted both qualitative and quantitative analyses using ResNet50 \cite{he2016deep}.

\noindent
\textbf{Qualitatively:} We employed t-distributed stochastic neighbor embedding (t-SNE) to visualize the logits distribution based on a fold of the WCE images (see Fig. 2). In Fig. 2(a), the distribution of the original WCE images, initially reduced to three dimensions using PCA, illustrates the challenge of high intra-class variance and inter-class similarity. Compared to Fig. 2(b), Fig. 2(d) exhibits a more compact intra-class distribution and a more diffuse inter-class distribution, highlighting the effectiveness of our proposed $\mathcal{L}_{DSCL}$. Additionally, Fig. 2(c) illustrates the distribution of $\mathcal{L}_{CE}$ using the zoomed-in WCE images, ruling out the effectiveness observed in Fig. 2(d) are attributed to the zoomed-in images.

\noindent
\textbf{Quantitatively:} Logits from the final layer of the network were utilized to calculate cosine similarity. Table 3 reveals that $\mathcal{L}_{DSCL}$ achieves higher intra-class similarity and lower inter-class similarity. SaliencyNet \cite{recasens2018learning} is conducted to assess the impact of zoomed-in WCE images on the model.

\section{CONCLUSION}
\label{sec:majhead}

In this paper, we propose a novel \textit{DSCL} approach to tackle inherent challenges posed by higher intra-class variance and inter-class similarities within the WCE domain. By utilizing saliency maps to zoom in on lesion regions, our method facilitates feature extraction, allowing the capture of rich and discriminative information within and across different classes in WCE images. Our extensive experimental results, conducted on a combination of two publicly available WCE datasets, demonstrate the effectiveness and superiority of our proposed method compared to other methods.

\vfill
\pagebreak


\bibliographystyle{IEEEtran}
\ninept
\bibliography{strings,refs}

\end{document}